\newif\ifieee
\IfFileExists{IEEEtran.cls}{\ieeetrue}{\ieeefalse}
\ifieee
  \documentclass[conference]{IEEEtran}
  \IEEEoverridecommandlockouts
\else
  \documentclass[10pt,twocolumn]{article}
  \usepackage[letterpaper,margin=0.7in]{geometry}
  \newcommand{\IEEEoverridecommandlockouts}{}

  \newenvironment{IEEEkeywords}{\par\small\noindent\textbf{Index Terms---}}{\par}
\fi
\usepackage{amsmath,amssymb,bm}
\usepackage{booktabs}
\usepackage{cite}
\usepackage{microtype}
\usepackage{url}

\title{Rethinking Image Processing for the Age of AI:\\
A Problem-First Framework for Scientific Progress} 
\date{}
\author{Guoping Qiu 
 \IEEEcompsocitemizethanks{\IEEEcompsocthanksitem G. Qiu is with the University of Nottingham Ningbo China, Ningbo, China, and the University of Nottingham, Nottingham, United Kingdom. E-mail: guoping.qiu@\{nottingham.edu.cn, nottingham.ac.uk\}
}
}

\begin{document}
\maketitle
\begin{abstract}
Modern AI has greatly expanded the capabilities of image processing. However, the ready availability of powerful models, public datasets, and benchmark leaderboards has also encouraged a model-first research pattern: researchers increasingly begin with an available architecture and optimize it on a public benchmark, rather than beginning with the underlying real-world imaging problem. This can produce impressive benchmark results without necessarily improving our understanding or solution of the real problem. This paper argues for a problem-first approach that distinguishes the physical imaging problem, solution principle, statistical estimator, and computational implementation, while clarifying what modern AI can achieve and which fundamental problems remain unsolved. Through case studies of super-resolution and low-light enhancement, we show how benchmark datasets may define tasks that differ substantially from the real-world problems they are intended to represent, and why performance improvements must be interpreted within the conditions under which they are obtained. We propose a six-stage workflow that places problem formulation, image acquisition, information-loss analysis, assumptions, ambiguity, and evaluation before model and dataset selection. The paper also proposes clearer standards for evidence, reproducibility, uncertainty, and claims of state-of-the-art performance. More fundamentally, it calls for a change in research culture and education so that future researchers learn to understand imaging problems deeply and use modern AI to achieve genuine scientific and technical advancement.
\end{abstract}


\section{Introduction}
Classical image processing designed operators: convolutional filters, transform-domain rules, interpolation kernels, Wiener estimators, and regularized inverses. Learning by example moved part of this design into data, including early neural restoration, learned inter-resolution prediction, exemplars, dictionaries, and learned priors \cite{wiener1949,tikhonov1963,qiu1999,freeman2002,aharon2006}. Deep networks now learn representations and mappings jointly and dominate restoration benchmarks \cite{dong2014,zhang2017,liang2021}.

This history is often presented as though modern learning methods have made the problems and principles of classical image processing less relevant. We argue for a more discriminating view. Four levels should not be conflated:
\begin{equation}
\boxed{\text{problem}\ne\text{principle}\ne\text{estimator}\ne\text{implementation}.}
\label{eq:fourlevels}
\end{equation}
A convolutional neural network (CNN) or transformer is an implementation family. Conditional-mean estimation is a statistical estimator. Bayesian inference or empirical risk minimization is a solution principle. These concepts describe how a solution is formulated, estimated, or implemented; they do not define the imaging problem itself. The imaging problem must instead be defined by the desired output, the physical acquisition process, the information lost or corrupted, and the intended application.

The practical concern is no longer only whether a network improves PSNR, perceptual quality, or a challenge ranking. A model can be optimal for the distribution defined by its paired-data construction and still be systematically wrong for deployment. Conversely, no finite ``real-world'' dataset can represent every camera, illumination, processing pipeline, scene, user preference, and future domain. The field therefore needs both a clearer understanding of what benchmark results establish and a systematic approach for developing reliable methods when training data cannot fully represent real-world applications.

This need is especially important for researchers entering the field. Open-source implementations, pretrained models, and public benchmark datasets have greatly accelerated experimentation and widened participation. At the same time, publication and evaluation systems naturally favour results that are rapid, measurable, and directly comparable. It is therefore rational for researchers to begin with an available or fashionable architecture and a public benchmark dataset, optimize the model for that dataset, and only afterwards ask what physical problem was solved. In that sequence, a small numerical gain may reflect architecture, parameter count, training schedule, augmentation, random variation, test-set familiarity, or tuning effort rather than a new understanding of image formation or recovery. Calling every such gain ``state of the art'' can unintentionally turn a conditional leaderboard result into a general scientific claim. Moreover, when code, data construction, hyperparameters, selection criteria, computational budget, and repeated-run variation are incompletely reported, the ordering itself may not be independently reproducible. These observations do not diminish the value of careful benchmark research. They explain why current research practices have developed and identify the evidence needed so that later research can build reliably on the findings.

A deeper concern is what this research pattern teaches. When the normal entry point into the field is an existing codebase, a fashionable model family, and a benchmark table, students may learn to regard the architecture as the intellectual centre of the problem. They can become skilled at changing blocks, losses, schedules, and hyperparameters without developing an equally strong understanding of optics, sensing, sampling, noise, dynamic range, colour formation, identifiability, prior information, or decision theory. The result is not a lack of technical ability; it is an imbalance between computational proficiency and scientific understanding. Such an imbalance makes it harder to recognize when a benchmark represents the wrong conditional problem, when an apparent improvement comes from a stronger prior rather than better recovery, or when an entirely different formulation is needed.

This concern is not confined to one institution or research tradition. In informal discussions with peers at major international conferences, universities, national laboratories, and technology companies, a recurring concern is that students, researchers, and new recruits increasingly begin by tuning models and often have a weaker grasp of the physical and inferential foundations of the task. These discussions are not a formal survey and should not be read as quantified evidence. They nevertheless point to a widely recognized educational challenge: the field must preserve the speed and power of modern AI while restoring disciplined study of the problems to which it is applied.

The central educational and methodological proposition of this paper is therefore simple: study the imaging problem before selecting the model. A researcher should understand how the observation was formed, what information survives, what has been lost, which assumptions or priors can legitimately restore it, what ambiguities remain, and what output the application actually requires. Only then should a benchmark, loss, architecture, and training protocol be chosen. This ordering leaves ample room for powerful modern AI, but makes model novelty serve a defined scientific question rather than substitute for one.

This paper makes eight contributions.
\begin{enumerate}
\item It develops a first-principles decomposition of imaging research into \emph{problem, solution principle, estimator, and implementation}, and a six-source account of the knowledge that resolves missing information. This provides a common language for designed filters, Bayesian/regularized reconstruction, example-based learning, deep networks, transformers, and generative priors.
\item It clarifies what deep learning and modern AI can achieve through representation learning, increased capacity, wider contextual modelling, optimization, and large empirical priors, while identifying fundamental problems that remain unresolved, including limits imposed by acquisition, sampling, non-identifiability, decision objectives, and dependence on the training distribution.
\item It formalizes benchmark--deployment mismatch and audits representative super-resolution (SR) and low-light image enhancement (LLIE) datasets. DIV2K-style digitally downsampled SR, RealSR/DRealSR capture rigs, LOL paired exposures, SID short/long raw exposures, and SICE selected HDR renderings are shown to define related but non-equivalent conditional tasks. It distinguishes insufficient spatial capture in SR, and limited dynamic range and low SNR in LLIE, from the convenient artificial transformations used to construct many benchmarks.
\item It turns the analysis into a six-stage problem-first workflow and applies every stage to SR and low-light enhancement, contrasting the resulting research design with representative CNN, transformer, GAN, blind-degradation, Retinex, burst-acquisition, inverse-rendering, null-space, and diffusion approaches. The workflow is also intended as a practical guide for students and researchers entering image processing.
\item It proposes a programme for progress under irreducible mismatch between training data and real-world applications: physically meaningful forward-model families, robust and worst-group evaluation, cross-dataset and cross-sensor testing, measurement-constrained adaptation, calibrated uncertainty, benchmark portfolios, deployment feedback, and explicitly scoped claims.
\item It articulates standards for reproducible comparison and a ladder separating numerical, comparative, mechanistic, generalization, and problem-progress claims, then translates the framework into guidance for research training, authors, reviewers, and benchmark organizers.
\item It identifies researcher education as part of the scientific problem. It proposes a training sequence in which image formation, information loss, assumptions, ambiguity, and evaluation precede model selection, so that future researchers learn not only how to optimize systems but how to formulate and advance imaging problems.
\item It proposes a community-level cultural change that broadens what counts as progress and distributes responsibility across researchers, supervisors, reviewers, editors, benchmark organizers, institutions, and industry. The aim is to make problem formulation, diagnostic evidence, robustness, and reproducibility as visible and rewardable as leaderboard performance.
\end{enumerate}

The claim is therefore not that ``nothing has changed,'' nor that modern AI, benchmarks, or current researchers are misguided. Deep learning can turn previously limited principles into powerful technologies, and benchmarks are indispensable for controlled comparison. The model-first pattern arises partly because the surrounding research system makes it efficient and rewardable. The contribution is to place models and benchmarks in the correct inferential hierarchy and to suggest how the community can broaden its incentives and evidence. By distinguishing improvement in architecture from improvement in problem formulation, data, estimator, or evidence base, the framework aims to help researchers design experiments that generalize more reliably, help reviewers and editors recognize diverse forms of rigorous contribution, and direct technical innovation toward the causes of imaging failure rather than benchmark-specific symptoms.

\section{Why Image Processing Needs Rethinking}
\subsection{Speed of experimentation versus depth of understanding}
Modern software has reduced the cost of entering image-processing research. A student can download a dataset, modify a network block, launch training, and obtain a complete table of standard metrics in a short time. This is an important democratization of research. It allows ideas to be tested rapidly and permits communities to build on shared implementations rather than repeatedly reconstructing basic infrastructure.

The same convenience, however, can invert the scientific order. A method may begin with the question ``where can this model be applied?'' rather than ``what prevents this image from containing the desired information?'' The benchmark then supplies the problem definition by default, the loss supplies the meaning of quality, and the architecture supplies the presumed novelty. Extensive training can make the resulting system highly optimized for this closed loop even when the loop omits important causes of the deployed failure.

Real scientific progress often moves more slowly. It may require characterizing a sensor, deriving a forward model, discovering which variables are confounded, constructing controlled data, deciding what constitutes a legitimate reference, and designing an evaluation that can falsify the central claim. These activities can yield less immediate numerical reward than replacing one backbone by another, yet they frequently determine whether the eventual technology works outside the laboratory. Patience is not an alternative to computational progress; it is what gives computation a meaningful target.

\subsection{Why the model-first pattern is rational}
The model-first pattern should not be attributed to a deficiency of individual researchers. It is a rational response to the environment in which research is conducted. Reusable implementations and pretrained weights lower the cost of model experiments. Standard datasets and metrics make results easy to compare. Leaderboards make rank visible. Short publication cycles favour projects whose outcomes can be produced and communicated quickly. Reviewers reasonably ask for comparison with established baselines, while editors and programme committees need evidence that can be assessed consistently across many submissions.

By contrast, problem-first work often has higher initial cost and less predictable output. It may require access to sensors, expertise across optics and statistics, new data collection, careful target construction, or experiments that reveal that a convenient formulation is inadequate. A better forward model, a negative result, or a diagnosis of non-identifiability may be scientifically important without immediately producing the highest score in a familiar table. Under current incentives, choosing the more measurable path is understandable.

The required cultural change must therefore make rigorous problem work more visible and more rewardable. It should broaden, not replace, the community's concept of contribution. Architectural advances and benchmark improvements remain valuable, but should sit alongside advances in problem formulation, acquisition modelling, dataset construction, uncertainty, evaluation, diagnosis, and deployment evidence. The goal is to encourage researchers to produce reliable advances that future work can build upon.

\subsection{The educational cost of a model-first paradigm}
Research practice and researcher education cannot be separated. Students learn what a field considers important from the work they are encouraged to reproduce, the questions asked in group meetings, the criteria used in paper review, and the achievements rewarded by leaderboards. If success is repeatedly presented as selecting a recent architecture and obtaining a higher score on an established public benchmark dataset, this procedure gradually becomes an implicit curriculum.

That curriculum can produce researchers who are highly competent in software engineering, optimization, and empirical experimentation but less prepared to ask foundational questions. They may know how to train a restoration transformer without being able to explain which spatial frequencies were lost; how to apply a diffusion prior without distinguishing plausible completion from measurement-supported recovery; or how to optimize low-light enhancement without analysing exposure, photon noise, clipping, illumination, and display rendering. The weakness is not the use of modern models. It is the absence of the problem knowledge needed to decide when, why, and under what conditions those models are appropriate.

This has consequences for innovation. Without a firm understanding of the physical and statistical problem, novelty is likely to be sought within the space already defined by the benchmark: another backbone, module, loss term, or training recipe. Fundamental progress often requires stepping outside that space---redefining the target, modelling a neglected cause, collecting a different measurement, exposing non-identifiability, or changing the evaluation. Researchers can only make that step when they understand the problem independently of the current solution machinery.

The required change is therefore both methodological and educational. Problem-first research should be taught as a habit of scientific reasoning: begin with the world and its measurements; determine what is known, lost, and ambiguous; justify the information introduced by a solution; define the intended decision; and then select the most appropriate model. Modern AI belongs fully within this curriculum, but as a powerful means of inference rather than the starting definition of the problem.

\subsection{Three meanings of progress}
The word \emph{progress} is commonly used for three different achievements:
\begin{enumerate}
\item \textbf{Benchmark progress:} improvement of a metric under a fixed dataset and protocol.
\item \textbf{Method progress:} a better estimator, representation, optimizer, or implementation under stated conditions.
\item \textbf{Problem progress:} better understanding or solution of the intended physical and decision problem over a meaningful deployment region.
\end{enumerate}
Benchmark progress can provide evidence for method progress, and method progress can contribute to problem progress. Neither implication is automatic. A 0.1 dB improvement may be valuable if achieved reproducibly under a demanding controlled comparison; it may be immaterial if obtained through additional supervision, compute, data, or test-specific tuning that is not acknowledged. Conversely, a method that sacrifices a fraction of a benchmark score may constitute greater problem progress if it handles unknown sensors, exposes uncertainty, preserves data consistency, or reveals a previously ignored failure mechanism.

\subsection{A hierarchy of scientific questions}
Before asking which architecture performs best, an imaging study should answer a hierarchy of questions:
\begin{enumerate}
\item What physical quantity or rendered image is desired, and for what use?
\item How was the observation formed, and which variables influenced it?
\item What information about the target remains in the observation?
\item What information has been destroyed or confounded?
\item Which additional assumptions, data, or priors are scientifically legitimate?
\item What output or decision is most appropriate for the intended application, and which types of error matter most?
\item Which computational method implements that action effectively?
\item What evidence would show that the method works beyond the development benchmark?
\end{enumerate}
This hierarchy does not demand a perfect physical model before learning can begin. It demands that approximation be explicit. A deliberately simplified model can be scientifically powerful because its domain of validity can be investigated; an implicit model hidden in a dataset is harder to interrogate.

\begin{table*}[t]
\centering
\caption{Four levels that should be reported separately in an image-processing study.}
\label{tab:four-levels}
\small
\begin{tabular}{p{2.5cm}p{4.0cm}p{4.1cm}p{3.4cm}}
\toprule
Level & Central question & Typical choices & Typical failure when conflated\\
\midrule
Problem & What latent quantity should be inferred from what observation, under which acquisition and use? & Sensor-to-radiance recovery, canonical rendering, perceptual enhancement, recognition-oriented preprocessing & A benchmark construction silently becomes the real-world problem\\
Principle & What knowledge makes the problem solvable? & Bayesian inference, regularization, empirical risk minimization, invariance, self-consistency & A generic learning principle is presented as a problem-specific explanation\\
Estimator & Which decision is optimal for the loss and uncertainty? & Conditional mean, median, MAP estimate, posterior sample, robust decision & Different outputs are ranked as if they estimated the same target\\
Implementation & How is the estimator represented and computed? & Filter, optimization, CNN, transformer, diffusion sampler, hybrid system & Architectural novelty is mistaken for new information or new physics\\
\bottomrule
\end{tabular}
\end{table*}

\section{The Information Problem}
A broad imaging model is
\begin{equation}
Y=\mathcal H(X;\eta)+N,
\label{eq:forward}
\end{equation}
where $X$ is the desired image, $Y$ the observation, $\mathcal H$ the acquisition or degradation process, $\eta$ its parameters, and $N$ disturbance. The linear case is
\begin{equation}
Y=HX+N.
\label{eq:linear}
\end{equation}
Restoration constructs
\begin{equation}
\widehat X=\delta(Y).
\label{eq:decision}
\end{equation}
But if
\begin{equation}
\mathcal H(X_1)=\mathcal H(X_2),\qquad X_1\ne X_2,
\label{eq:nonidentifiable}
\end{equation}
the observation cannot distinguish the two latent images. Depth, attention, or parameter count cannot restore uniqueness; an estimator must inject additional knowledge.

We distinguish six sources:
\begin{equation}
\mathcal K=\{K_m,K_p,K_a,K_i,K_e,K_r\},
\label{eq:knowledge-set}
\end{equation}
denoting information retained in the measurement, acquisition physics, analytic assumptions, internal image statistics, external examples, and representation/architecture bias, respectively. Conceptually,
\begin{equation}
\widehat X=F(Y;K_p,K_a,K_i,K_e,K_r).
\label{eq:knowledge-estimator}
\end{equation}
Classical interpolation emphasizes $K_a$; model-based inversion combines $K_p$ and $K_a$; example-based methods use $K_e$; zero-shot super-resolution uses $K_i$ \cite{shocher2018}; and Deep Image Prior demonstrates the utility of $K_r$ without external training \cite{ulyanov2018}. The useful question is therefore not simply whether a method learns, but which source resolves which ambiguity.

A method may use several of the information sources in \eqref{eq:knowledge-set} together. Considering each source separately, however, makes it easier to identify why the method may fail. Possible causes include an inaccurate model of the imaging process, insufficient information in the observed image, differences between the training and deployment data, and learned preferences for structures that are not present in the actual scene. 

\subsection{Identifiability, stability, and support}
Three questions should be separated. \emph{Identifiability} asks whether distinct latent images can produce the same observation. \emph{Stability} asks whether small perturbations in the observation or model can cause large changes in the reconstruction. \emph{Support} asks whether the deployed observation lies in a region for which the estimator has relevant knowledge. Deep capacity can approximate a complicated inverse on the training support, but it cannot, by itself, recover information that was not preserved during image acquisition, make an unstable inverse stable without regularization, or guarantee behaviour outside that support.

If $\mathcal H$ is differentiable locally, small changes satisfy approximately
\begin{equation}
\Delta Y\simeq J_{\mathcal H}(X)\Delta X+\Delta N.
\label{eq:local-forward}
\end{equation}
Small singular values of $J_{\mathcal H}$ identify directions in image space that are weakly observed; its null directions —changes in the underlying image that the acquisition process cannot detect — are not observed at all. A network may output visually convincing content in these directions because the training distribution makes some contents more probable than others. That is successful prior-based inference when the task permits plausibility, but it is not direct recovery from the measurement.

\subsection{Evidence-supported and prior-supported content}
The distinction between evidence and prior is continuous rather than binary. For a reconstructed feature $\phi(X)$, one may ask how sensitive it is to the measurement, the forward model, and the prior. A feature that persists under plausible changes of prior but changes predictably with the observation is relatively evidence-supported. A feature that varies strongly across equally plausible priors or posterior samples is prior-supported. This suggests practical diagnostics: perturb the assumed degradation, vary the learned prior or sampling seed, enforce re-measurement consistency, and visualize uncertainty in weakly observed regions.

The word \emph{hallucination} is sometimes used for any generated detail. A more precise account distinguishes: (i) plausible completion consistent with the intended task; (ii) prior-dominated detail that is not identifiable but is honestly represented as uncertain; and (iii) confidently asserted content inconsistent with the measurement or application. The scientific concern is principally the third category, and the second category when it is presented as factual recovery.

\subsection{The target is part of the problem}
Many image-processing tasks do not possess one natural ground truth. A long exposure is not the unique truth corresponding to a short exposure when objects move. A fused or tone-mapped HDR image is a rendering decision. A high-resolution capture at another focal length is not the latent array that would have been recorded by an ideal high-resolution version of the original sensor. A normally exposed photograph changes exposure, noise, clipping, colour processing, and sometimes illumination relative to a low-light photograph.

Thus the target variable $X$ must be defined, not merely supplied. Is it scene radiance, reflectance, a canonical camera rendering, a preferred photograph, or an input optimized for a downstream machine? Ambiguous target construction changes $p(X|Y)$ and therefore changes the optimal estimator. Dataset documentation is consequently part of problem formulation rather than an administrative detail.

\section{One Statistical Bridge Across Three Eras}
Classical regularized inversion writes
\begin{equation}
\widehat X=\arg\min_X D(Y,\mathcal H(X))+\lambda R(X),
\label{eq:regularization}
\end{equation}
while Bayesian inference uses
\begin{equation}
p(X|Y)\propto p(Y|X)p(X).
\label{eq:bayes}
\end{equation}
Given paired examples $\mathcal D=\{(Y_i,X_i)\}_{i=1}^n$, a learned implementation uses
\begin{equation}
\theta^*=\arg\min_\theta\sum_i L(F_\theta(Y_i),X_i).
\label{eq:erm}
\end{equation}
These forms relocate knowledge---from explicit $R$ or $p(X)$ into examples and parameters---but need not change the target decision rule. For loss $L(X,a)$, the Bayes action is
\begin{equation}
\delta^*(y)=\arg\min_a\mathbb E[L(X,a)|Y=y].
\label{eq:bayes-action}
\end{equation}
Under squared error,
\begin{equation}
\delta^*(y)=\mathbb E[X|Y=y].
\label{eq:mmse}
\end{equation}
For jointly Gaussian variables this is linear and underlies Wiener estimation \cite{wiener1949}. A shallow network, CNN, or transformer may approximate progressively richer versions of the same conditional mean. Thus deep learning changed representation learning, capacity, receptive field, optimization, and the scale of empirical priors. It did not repeal sampling, information loss, identifiability, the dependence of an optimum on its loss, or the dependence of a learned estimator on its data distribution.

Generative methods do expand the available decision rules. Beyond the conditional mean, one may choose
\begin{equation}
\widehat X_{\rm MAP}=\arg\max_Xp(X|Y)
\label{eq:map}
\end{equation}
or sample
\begin{equation}
\widetilde X\sim p(X|Y).
\label{eq:sampling}
\end{equation}
The latter can represent multimodality, but plausible detail is not automatically measurement-supported detail \cite{blau2018,ho2020}.

\begin{table*}[t]
\centering
\caption{A problem-first comparison of three broad image-processing eras. The boundaries are porous; these are dominant tendencies rather than exclusive properties.}
\label{tab:eras}
\setlength{\tabcolsep}{5pt}
\small
\begin{tabular}{p{2.2cm}p{3.9cm}p{3.9cm}p{3.9cm}}
\toprule
Dimension & Designed/model-based processing & Learning by example & Deep and generative learning\\
\midrule
Representation & Usually designed; assumptions visible in filters, transforms, and regularizers & Designed features with learned predictors, dictionaries, exemplars, or priors & Representation and processing learned jointly at high capacity\\
Primary knowledge & Physics and analytic assumptions ($K_p,K_a$) & Internal or external examples ($K_i,K_e$), often with explicit structure & Large external priors, architecture bias, and increasingly physics/internal adaptation\\
Typical inference & Direct filtering or explicit optimization & Regression, lookup, retrieval, or sparse inference & End-to-end regression, unfolding, adversarial inference, or posterior sampling\\
Main strength & Transparency, data efficiency, predictable constraints & Empirical flexibility with moderate interpretability & Capacity, context, transfer, and complex or multimodal conditionals\\
Characteristic risk & Model bias and restrictive priors & Limited examples or representation capacity & Distribution shift, opaque priors, and convincing unsupported detail\\
What remains invariant & \multicolumn{3}{p{11.7cm}}{Observation formation, information loss, identifiability, loss-dependent decisions, and the need to justify prior information}\\
\bottomrule
\end{tabular}
\end{table*}

Table~\ref{tab:eras} shows why ``classical versus learning'' is incomplete. Dictionary methods learn representations but retain explicit inference; plug-and-play methods insert learned denoisers into model-based solvers; and diffusion restoration can combine a likelihood with a learned score prior \cite{venkatakrishnan2013,romano2017,kawar2022}. A better taxonomy asks what is known, what is learned, what decision is made, and how it is computed.

\section{What Modern AI Changes and What It Does Not}
\subsection{Changes that are scientifically consequential}
Modern AI has changed image processing in at least six substantive ways. First, learned hierarchical representations reduce dependence on manually designed features. Second, large receptive fields and attention allow nonlocal and semantic context to influence local reconstruction. Third, scale permits empirical priors to encode far more diverse image statistics than classical parametric models. Fourth, amortized inference moves computation from repeated per-image optimization into training, enabling fast deployment. Fifth, generative models can represent multimodal conditional distributions rather than a single point estimate. Sixth, pretraining and adaptation permit knowledge to be transferred across tasks and domains.

These are not cosmetic implementation changes. They can alter which priors are available, which estimator can be approximated, and which problems are computationally feasible. A problem-first view therefore does not imply a return to hand-designed filters. It asks that these new capabilities be located accurately in the inference chain.

\subsection{Persistent constraints}
Several constraints remain. The sensor still samples a physical field; saturation still discards magnitude; downsampling still removes or aliases frequencies; noise still reduces information; and multiple latent scenes can still explain one observation. The optimum remains conditional on a loss and distribution. A generative model can produce realistic textures, but realism is a property of a distribution or observer response, not proof that the particular texture occurred in the measured scene.

Likewise, scale does not make training data neutral. Large pretraining datasets reflect the capture devices and processing conventions used to create them, as well as uneven representation of different cultures and visual content, licensing restrictions, and data-selection processes. 

\subsection{Hybridization is often the natural consequence}
The problem-first framework frequently leads to hybrids rather than a choice between physics and learning. Known acquisition can enter through a likelihood or data-consistency term; unknown parameters can be estimated; a learned prior can supply image statistics; internal adaptation can use the test observation; and uncertainty can be represented through posterior samples or calibrated intervals. The resulting system may still use a transformer or diffusion model, but its role is explicit.

This also clarifies interpretability. A completely transparent small model is not automatically correct, and a large network is not automatically uninterpretable at the scientific level. Even when individual features are opaque, the system can remain scientifically interpretable if the target, acquisition family, information sources, loss, consistency constraints, and validated domain are explicit.

\section{Benchmarks as Scientific Instruments}
\subsection{What a benchmark legitimately provides}
A benchmark freezes some variables so that methods can be compared. It may provide common data, a reference construction, an evaluation metric, and a public protocol. These controls reduce experimental degrees of freedom and allow a community to accumulate results. Synthetic benchmarks additionally permit exact pairing and controlled perturbation; captured benchmarks incorporate effects that a simulator may miss. Both are indispensable.

But a benchmark is an instrument with a transfer function. Its selection of scenes, sensors, targets, degradations, preprocessing, and metrics determines what it is sensitive to and what it ignores. A score is evidence about performance under that instrument. Generalization to a broader claim requires additional argument and experiments.

Let $B$ denote the benchmark construction, including its acquisition or synthesis process, sampling of scenes, target definition, and metric. A reported score is more precisely
\begin{equation}
S(F;B,\Pi,T,C),
\label{eq:score-context}
\end{equation}
where $\Pi$ is the evaluation protocol, $T$ the model-selection and tuning procedure, and $C$ the allowed resources. Omitting $\Pi,T,$ or $C$ makes apparently identical comparisons potentially unequal.

\subsection{A taxonomy of benchmark mismatch}
Benchmark--deployment mismatch has several distinct forms:
\begin{enumerate}
\item \textbf{Forward-process mismatch:} synthetic blur, noise, exposure, or compression differs from acquisition.
\item \textbf{Scene and content mismatch:} subjects, textures, motion, geography, or semantics differ.
\item \textbf{Sensor and pipeline mismatch:} optics, CFA, ISP, colour response, bit depth, and codec differ.
\item \textbf{Target mismatch:} the reference image represents a different exposure, rendering, viewpoint, or preference.
\item \textbf{Loss and use mismatch:} the optimized metric does not reflect human judgement, fidelity, safety, or downstream performance.
\item \textbf{Protocol mismatch:} preprocessing, crop conventions, colour spaces, test adaptation, or external data differ.
\end{enumerate}
These mismatches should not be collapsed into a generic statement that ``real data are different.'' Each suggests different diagnostics and remedies.

\subsection{Benchmark saturation and adaptive overfitting}
A test set may be nominally held out while influencing research indirectly through repeated publication, leaderboard feedback, visual inspection, and community-wide tuning. Over time, architectural and training choices become adapted to its idiosyncrasies. This is not necessarily misconduct; it is a predictable consequence of repeated selection. Studies using newly collected ImageNet-like test sets illustrate how performance can change under even carefully reconstructed data collection \cite{recht2019imagenet}. The scientific response is to maintain hidden or refreshed tests, use multiple independent benchmarks, report cross-dataset transfer, and resist interpreting tiny mature-benchmark gains as broad progress without corroboration.

\begin{table*}[t]
\centering
\caption{A portfolio of evidence for claims about an imaging method.}
\label{tab:evidence-portfolio}
\small
\begin{tabular}{p{3.0cm}p{5.2cm}p{5.3cm}}
\toprule
Evidence component & What it establishes & What it does not establish alone\\
\midrule
Matched synthetic benchmark & Controlled accuracy for a known degradation and reference & Robustness to unknown acquisition or real sensor pipelines\\
Captured paired benchmark & Performance for a specific capture and registration protocol & Universal real-world generalization or unique ground truth\\
Cross-dataset/sensor transfer & Sensitivity to dataset and device shift & Coverage of all deployments\\
Forward-model stress test & Behaviour under controlled physical parameter changes & Adequacy of the chosen model family\\
Re-degradation consistency & Compatibility of output with observation under an assumed model & Truth of prior-dominated null-space detail\\
Unpaired in-the-wild testing & Qualitative failure modes and deployment diversity & Full-reference fidelity\\
Human/downstream evaluation & Utility for a specified observer or task & Physical faithfulness for other uses\\
Repeated runs and ablations & Stability and attribution under the reported protocol & Generalization beyond that protocol\\
\bottomrule
\end{tabular}
\end{table*}

\section{State of the Art, Reproducibility, and Scientific Claims}
\subsection{SOTA is a relation, not a property}
No method is simply ``state of the art.'' It is state of the art with respect to a dataset, split, metric, preprocessing convention, resource envelope, and date. Different losses can reverse rankings; confidence intervals can overlap; and increased data or compute may account for the difference. The phrase is scientifically useful only when these conditions are visible.

We propose a claim ladder:
\begin{enumerate}
\item \emph{Numerical claim:} the submitted outputs obtain a stated score.
\item \emph{Comparative claim:} the method outperforms specified baselines under matched conditions.
\item \emph{Mechanistic claim:} evidence attributes the gain to the proposed idea rather than confounds.
\item \emph{Generalization claim:} the gain survives relevant shifts and independent tests.
\item \emph{Problem-progress claim:} the method improves the intended real task or advances its scientific understanding.
\end{enumerate}
Each step requires new evidence. A leaderboard normally establishes at most the first two.

\subsection{Reproducibility has several levels}
\emph{Computational reproducibility} asks whether the same code and environment recover the result. \emph{Experimental reproducibility} asks whether an independent implementation under the described protocol reaches the same conclusion. \emph{Inferential reproducibility} asks whether reasonable changes of dataset, seed, metric, or analysis preserve the claimed ordering. A downloadable checkpoint can satisfy the first while leaving the latter two unanswered.

For mature benchmarks where differences are small, papers should report enough information to reconstruct data generation, preprocessing, training schedules, model selection, test-time processing, external data, and compute. Repeated runs should be reported when randomness is comparable with the claimed gain. Baselines should receive a fair tuning budget and the comparison should separate the proposed idea from extra parameters, pretraining, supervision, or inference cost. These requirements align with broader machine-learning reproducibility practice, which treats transparent experimental workflows as part of the mechanism by which findings become accessible and reliable \cite{pineau2021reproducibility}.

\subsection{Negative and Diagnostic Results}

The current research environment rewards a new maximum on a benchmark more visibly than a careful diagnosis of failure. This can make students and early-career researchers reluctant to report negative results, particularly when they believe that a paper without a state-of-the-art score is unlikely to be accepted. Consequently, unsuccessful experiments are often omitted, common failures are repeatedly rediscovered by different groups, and research effort remains concentrated on small improvements within established settings. This creates an incomplete scientific record and can delay recognition that the current formulation, data, or evaluation protocol is inadequate.

A negative result, however, is not simply the absence of an improvement. A scientifically valuable negative result tests a clearly stated hypothesis under appropriate and carefully controlled conditions. It verifies that the observed failure is not caused by an implementation error, inadequate optimization, an unfair baseline, or insufficient experimental effort. It then establishes where the failure occurs, whether it is reproducible across methods and datasets, and which assumptions are responsible. For example, showing that several substantially different architectures fail under the same change in sampling, sensor, illumination, or noise conditions can demonstrate that the limitation lies in the problem formulation or available information rather than in a particular network design.

Diagnostic results can be especially valuable in image processing because they reveal the boundary between conditions under which a method is reliable and conditions under which it is not. Sensitivity analysis can determine how performance changes as blur, noise, exposure, sampling, compression, or distribution mismatch increases. Ablation and controlled interventions can identify which source of information supports a method and when that source becomes unreliable. Cross-dataset and cross-sensor experiments can reveal whether an apparent advance is tied to one benchmark construction. Such findings map the validated operating region of a method and indicate where additional measurements, better physical models, different priors, or new evaluation protocols are needed.

Young researchers should therefore be encouraged to frame a well-supported negative result as a positive contribution to knowledge. A strong paper of this kind should state the original expectation, describe the controls used to test it, report the conditions under which it fails, provide evidence for the cause of failure, and explain the consequences for future research. Where possible, it should release the experimental protocol, failed cases, and relevant code or data so that other researchers can verify the conclusion and avoid repeating the same unsuccessful path. The contribution is not that ``the method did not work,'' but that the study has established why it did not work, under what conditions it fails, and what must change for further progress to become possible.

Reviewers and editors also have an important role. A paper should not be rejected merely because it does not report a new highest benchmark score. It should instead be judged according to whether the question is important, the experiments are rigorous, the negative conclusion is reproducible, plausible alternative explanations have been examined, and the findings change the community's understanding or future research decisions. This does not lower the standard for publication. On the contrary, a convincing negative result often requires stronger controls and more careful reasoning than a small positive improvement.

A problem-first research culture should therefore recognise calibrated negative results, sensitivity analyses, failure taxonomies, and mappings of validated operating conditions as legitimate scientific contributions. These studies reduce duplicated effort, prevent unsupported claims from becoming established assumptions, and identify where new measurements, formulations, or scientific ideas are required. Making such work visible and publishable would also allow young researchers to investigate difficult and uncertain questions without feeling that only a favourable numerical outcome can justify the effort.


\section{Case Study I: Super-Resolution and the Sampling Deficit}
Super-resolution is fundamentally a problem of insufficient spatial measurement. A camera does not observe an HR array and subsequently apply the digital resize operation used by a benchmark. Scene radiance is filtered by the optics, integrated over finite pixel areas, sampled on a sensor lattice, corrupted by noise, and processed by a camera pipeline. A useful abstraction is
\begin{equation}
Y=\mathcal P\!\left[D_s(KX)+N\right],
\label{eq:sr}
\end{equation}
where $K$ represents optical blur and sensor integration, $D_s$ represents spatial sampling, $N$ is measurement noise, and $\mathcal P$ contains operations such as mosaicing, demosaicing, sharpening, quantization, and compression. When scene frequencies exceed the effective passband and sampling limit, they are attenuated, removed, or aliased. The observation therefore contains too few reliable independent measurements to determine the desired HR image.

Let $A$ denote the resulting linearized measurement operator. Its null space
\begin{equation}
\mathcal N(A)=\{v:Av=0\}
\label{eq:nullspace}
\end{equation}
implies
\begin{equation}
A(X+v)=AX,\qquad v\in\mathcal N(A).
\label{eq:null-ambiguity}
\end{equation}
This non-identifiability, rather than visible pixelation itself, is the root inference problem. Conceptually,
\begin{equation}
X=A^\dagger Y+(I-A^\dagger A)Z,
\label{eq:sr-decomposition}
\end{equation}
where $A^\dagger Y$ is constrained by the measurement and $(I-A^\dagger A)Z$ lies in the unobserved null space. A single-image algorithm can estimate the latter only by importing assumptions or prior information. It cannot establish that prior-supported detail is the uniquely recovered scene.

\subsection{Two physical causes obscured by one benchmark label}
Practical SR couples two causes. The first is \emph{insufficient sampling}: the optics--sensor system has not acquired enough spatial measurements to represent the desired bandwidth. The second is \emph{limited measurement quality}: optical attenuation, pixel integration, noise, CFA sampling, demosaicing, compression, and camera processing reduce the reliability or accessibility of the samples that remain. These causes interact. Aliasing may preserve high-frequency energy in displaced form, while optical low-pass filtering can remove it; noise can make nominally observed frequencies unusable; and demosaicing can mix colour reconstruction with spatial-resolution loss.

A fixed bicubic reduction (or any other sub-sampling operation) does not reproduce this causal process. It begins with an already captured HR image and digitally constructs
\begin{equation*}
Y_b=D_s(K_bX),
\end{equation*}
where $K_b$ and $D_s$ are prescribed by the benchmark. This creates a clean and reproducible conditional task, but it should not be confused with the original problem of acquiring insufficient, noisy, possibly aliased samples through an unknown optical--sensor pipeline. Similar visual appearance does not imply identical missing information.

\subsection{What the recent literature addresses---and what it leaves implicit}
Classical interpolation fills missing samples using analytic smoothness; reconstruction methods impose explicit priors; external learning estimates LR--HR relations from datasets; and internal learning estimates them from recurrence within the test image. Early progressively predictive learning transferred a relation learned between known coarse levels to an unknown finer level \cite{qiu1999}; example-based SR retained external LR--HR exemplars \cite{freeman2002}; ZSSR trained an image-specific CNN from the test image \cite{shocher2018}; and modern CNNs scale external conditional regression \cite{dong2014}. All address the ambiguity in \eqref{eq:null-ambiguity}, but obtain and encode the missing information differently.

Important recent exceptions begin closer to the root cause. Handheld multi-frame SR exploits natural subpixel motion to acquire complementary RAW samples, increasing both spatial information and SNR \cite{wronski2019}; deep burst SR learns alignment and fusion of such multiple measurements \cite{bhat2021burst}. Null-space learning explicitly preserves the measurement-determined component and learns only the unobserved component from a GAN prior \cite{wang2023null}; unfolding methods derive network stages from data fidelity and an assumed degradation operator \cite{zhang2020usrnet}. These works exemplify different problem-first responses: collect more evidence, separate evidence from prior, or make the forward operator govern inference.

The main benchmark-driven SISR stream, however, commonly begins one level later. An HR public dataset is selected, a fixed degradation generates paired LR inputs, a CNN, transformer, GAN, or diffusion architecture is chosen, and training optimizes reconstruction or perceptual scores. Richer blind-degradation simulations improve transfer, and stronger generative priors improve visual plausibility, but neither changes the central fact that one LR observation does not determine its null-space content. A method can therefore respond impressively to the appearance of low resolution without directly addressing why the relevant information was not acquired.

This is not a claim that learned SISR lacks value or that every recent paper is model-first. It is a distinction between three kinds of advance: a better estimator for a prescribed benchmark conditional; a better prior for plausible synthesis; and better acquisition or inference for the underlying sampling problem. These contributions should not be collapsed into one ranking. A physically meaningful SR study should say which component is supported by measurements, which is imposed by prior information, and whether its objective is faithful reconstruction, blind camera restoration, cross-device translation, or plausible generation.

The epistemic distinction can be summarized as
\begin{equation}
\widehat X=X_{\rm evidence}+X_{\rm prior},
\label{eq:evidence-prior}
\end{equation}
where the decomposition need not be computationally unique. In scientific or medical imaging, reporting which structures are prior-dominated may matter as much as visual sharpness.

Thus, higher PSNR under bicubic degradation demonstrates a better estimator for that benchmark, not necessarily recovery of real sensor detail; perceptual sharpness may also trade against fidelity. Evaluation should align the acquisition model, loss-induced estimator, and application, using realistic degradations, data consistency, and uncertainty rather than one average score.

\paragraph{Benchmark mismatch in SR.}
The conventional SR benchmark is built from HR photographs rather than native LR acquisition. DIV2K provides 800/100/100 HR training/validation/test images, with LR inputs commonly generated by known bicubic downsampling \cite{agustsson2017div2k}; Set5, Set14, BSD100, Urban100, and Manga109 are tested similarly. This gives exact registration and an unambiguous target, but defines $p_{\rm bench}(Y|X)$ by the imposed kernel, omitting or simplifying optics, sensor sampling, demosaicing, signal-dependent noise, sharpening, compression, spatial variation, and processing order. It therefore benchmarks a matched simulator, not blind SR for an arbitrary camera.

Captured benchmarks reduce but do not remove the gap. RealSR changes focal length on two cameras and registers paired static scenes \cite{cai2019realsr}; DRealSR broadens real scenes and degradations \cite{wei2020drealsr}. They contain genuine optical/camera effects, but the HR target is a second capture, so focal setting, alignment/resampling, exposure and ISP differences enter the pair. Static scenes and limited rigs narrow the distribution: these are valuable real-pair benchmarks, not universal ground truth for real-world SR.

\subsection{Four different tasks called super-resolution}
At least four tasks are commonly grouped under one name. \emph{Known-degradation reconstruction} estimates an HR image under a specified $D,K,$ and noise model. \emph{Blind restoration} must infer or marginalize unknown degradation. \emph{Cross-device enhancement} learns a mapping defined by a particular paired capture rig. \emph{Generative upsampling} seeks a plausible high-resolution rendering conditioned on a low-resolution input. These tasks overlap, but they have different notions of correctness. PSNR against a bicubic-derived reference is appropriate for the first; perceptual preference may be central to the fourth; neither alone establishes faithful recovery from an unknown camera.

The distinction becomes more important as the prior becomes stronger. A diffusion or large generative prior can produce semantically convincing text, faces, vegetation, and masonry. If those patterns are weakly constrained by $Y$, the output may be useful for visual media while being unsuitable for evidence preservation, measurement, medicine, or remote sensing. The method should state which use it serves and expose prior dependence when factual fidelity matters.

\subsection{A stronger experimental design for SR}
A problem-first SR experiment would begin by deciding whether the information deficit should be reduced through acquisition or managed through inference. Where possible, multiple shifted frames, alternative sensor layouts, optical coding, or other complementary measurements should be considered before asking a single network to synthesize all missing detail. For a single observation, the study would begin with a deployment envelope rather than one degradation. It would calibrate or bound optical modulation, pixel integration, sampling and aliasing, noise, mosaicing, compression, and processing order; reserve combinations outside the training support; and include native low-resolution images not generated from the target. The report would separate:
\begin{enumerate}
\item accuracy on the matched simulator;
\item robustness to forward-parameter changes;
\item transfer to independent sensors and pipelines;
\item data consistency after re-applying plausible degradations;
\item perceptual utility where the task permits generative detail; and
\item uncertainty, prior sensitivity, or solution diversity in the null space; and
\item gains due to genuinely additional measurements from gains due to a stronger learned prior.
\end{enumerate}
This design can still identify a superior transformer or diffusion implementation, but its conclusion is stronger because the architecture is evaluated as one component of a defined inverse problem.

\section{Case Study II: Low-Light Enhancement and the Proxy Problem}
A common synthetic pair darkens a normally exposed image by a nonlinear curve and then adds noise, for example
\begin{equation*}
Y_s=Q\!\left[\mathcal G_\gamma(\alpha X)+N_s\right],
\qquad 0<\alpha<1,
\end{equation*}
where $\mathcal G_\gamma$ is a gamma or related tone curve. This is more realistic than simple intensity scaling because it can imitate dark appearance, nonlinear camera response, quantization, and noise. Nevertheless, it still starts from a well-exposed $X$ in which scene structure, colour, and shadow detail have already been captured. The transformation changes those recorded values; it does not reproduce why the values were difficult to record in the first place.

Real low-light formation is better described at the radiance and acquisition levels. Conceptually, let scene radiance be generated by reflectance $R$ under spatially varying illumination $L$ and recorded by a camera operator $\mathcal C$:
\begin{equation*}
Y_r=\mathcal C\!\left(R\odot L;t,\theta\right)+N,
\end{equation*}
where $t$ is exposure and $\theta$ collects aperture, sensor response, clipping, quantization, and in-camera processing. Photon scarcity and read noise reduce shadow SNR, while the coexistence of very dark and bright regions can exceed the usable dynamic range. Clipping at either end and coarse quantization are many-to-one operations: some information was never captured, rather than merely hidden by a gamma curve.

Low-light enhancement is therefore governed by two coupled root causes: \emph{insufficient dynamic range} and \emph{insufficient signal-to-noise ratio}. Wu \emph{et al.} formulate this distinction explicitly and address the former through a simplified inverse-rendering model \cite{wu2025inverse}. Their method recovers a simplified scene representation and re-renders it under a canonical light source, shifting the objective from inverting an arbitrary darkening curve to correcting image formation from first principles. In conceptual form,
\begin{equation*}
Y_r\ \longrightarrow\ (\widehat R,\widehat L)
\ \longrightarrow\
\widehat X_c=\mathcal C(\widehat R\odot L_c;t_c,\theta_c),
\end{equation*}
where $L_c$ denotes canonical illumination. This view separates scene reflectance from illumination and asks how the scene should be rendered, rather than assuming that a unique normal-light image can be recovered by applying the inverse of the synthetic gamma function.

\subsection{Two physical causes obscured by one benchmark label}
The HDR and low-SNR causes should be distinguished because they destroy information in different ways and therefore require different solution principles. Consider first an evenly illuminated scene recorded at a weak but spatially uniform light level. If the response is unsaturated, sufficiently sampled, and has adequate SNR, much of the visible darkness can in principle be corrected by exposure compensation and an appropriate rendering transform. Darkness by itself is not the difficult inverse problem. The difficulty begins when too few photons are collected: shot noise becomes large relative to signal, read noise and quantization become consequential, and amplification raises noise together with the desired signal. No tone curve or larger network can recreate measurement evidence that the sensor did not record.

The HDR cause is different. In many night, indoor, backlit, and unevenly illuminated scenes, bright light sources or well-lit surfaces coexist with deep shadows. A single exposure cannot place the entire radiance range inside the sensor's usable interval. Exposing for highlights drives shadows toward the noise floor; exposing for shadows clips highlights. The problem is therefore not adequately described as ``make a dark image brighter.'' It is to infer and render a scene whose spatial radiance range exceeds what the particular exposure and sensor can faithfully record. Illumination, reflectance, exposure, saturation, sensor noise, camera response, and the desired display rendering are coupled.

These causes also interact. A high-dynamic-range scene forces an exposure compromise that often creates low SNR in the dark regions; local brightening then exposes the noise produced by that compromise. Conversely, a globally photon-limited scene may have modest scene dynamic range but remain difficult because all measurements are uncertain. A scientifically defined LLIE task should therefore diagnose at least two variables,
\begin{equation*}
\underbrace{\rho_{\rm scene}/\rho_{\rm usable}}_{\text{dynamic-range pressure}}
\quad\text{and}\quad
\underbrace{\operatorname{SNR}(u)}_{\text{spatial measurement reliability}},
\end{equation*}
rather than treating ``low light'' as one scalar degradation strength. Here $\rho_{\rm scene}$ is the scene-radiance range, $\rho_{\rm usable}$ is the reliably recordable sensor range under the selected exposure, and $\operatorname{SNR}(u)$ varies spatially with the collected signal and sensor characteristics.

\subsection{What the recent literature addresses---and what it leaves implicit}
Important exceptions address parts of this diagnosis. RAW short-to-long-exposure learning models genuine photon and read noise more faithfully than gamma-darkened RGB pairs \cite{chen2018sid}; SNR-aware enhancement explicitly changes processing according to spatial measurement reliability \cite{xu2022snr}; Retinex-based methods represent illumination variation \cite{wei2018retinex,cai2023retinexformer}; and simplified inverse rendering treats the illumination and dynamic-range cause at the image-formation level \cite{wu2025inverse}. These works demonstrate that modern learning can be made substantially more effective when its role follows a physical diagnosis.

The broader literature, however, often begins one level later. A public dataset supplies low/normal pairs; its reference image is accepted as the target; a CNN, transformer, GAN, diffusion model, state-space model, or colour representation is selected; and the method is optimized for benchmark fidelity or perceptual quality. Noise suppression, illumination estimation, colour correction, and detail recovery are then introduced as network modules, frequently without first asking which failures arise from HDR exposure compromise, which arise from photon-limited measurement, which information is clipped or below the noise floor, and which output attributes are rendering preferences rather than recoverable scene properties. The method may respond successfully to visible symptoms while leaving the causal task under-specified.

It is not that all recent LLIE work ignores physics, nor that architectural innovation is unimportant. Rather, the main benchmark-driven stream rarely uses the two root causes to determine the target representation, required measurements, limits of recovery, training distribution, architecture, and evaluation protocol as one coherent chain. Consequently, methods addressing different conditionals---gamma inversion, paired-exposure translation, RAW photon-limited restoration, illumination correction, HDR reconstruction, or preferred tone rendering---are often compared in a common table as though they solved the same problem.

A problem-first study would reverse this order. It would first determine whether the deployment regime is dominated by dynamic-range pressure, low SNR, or both. It would then decide whether the scientifically appropriate intervention is improved acquisition (longer or multiple exposures, burst fusion, larger well capacity, RAW sensing, or HDR/event measurements), a calibrated noise estimator, inverse rendering, display tone mapping, or a combination. Only after those choices would it select the learned architecture. This diagnosis also changes evaluation: HDR-dominated cases require highlight/shadow information and rendering assessments, whereas SNR-dominated cases require signal-dependent noise, detail confidence, and cross-sensor tests. A single average PSNR on a mixed benchmark cannot reveal which root cause has actually been addressed.

Consequently,
\begin{equation*}
p_s(Y|X)\ne p_r(Y|X)
\end{equation*}
is not merely a difference in noise level. The synthetic and real processes can differ in their latent variables, dynamic-range constraints, clipping, and the information preserved in shadows and highlights. They therefore define different conditional inference problems.

Let the MSE-optimal estimators under training and deployment distributions be
\begin{equation*}
F^*_{\rm train}(y)=\mathbb E_{\rm train}[X|Y=y],\quad
F^*_{\rm real}(y)=\mathbb E_{\rm real}[X|Y=y].
\end{equation*}
Then 
\begin{equation}
\boxed{p_{\rm train}(X|Y)\ne p_{\rm real}(X|Y)
\ \Rightarrow\ F^*_{\rm train}\ne F^*_{\rm real}.}
\label{eq:wrong-problem}
\end{equation}
Equation~\eqref{eq:wrong-problem} formalizes why unlimited capacity and perfect optimization cannot rescue a benchmark whose paired-data mechanism defines the wrong conditional task. A learner may become excellent at undoing gamma and injected noise while failing on illumination variation, shadow SNR, saturation, colour shifts, and real camera processing. Synthetic data remain valuable when their rendering and sensor model preserve the relevant physics; the issue is not simulation itself but whether it captures the causes of the deployed low-light problem \cite{wei2018retinex,plotz2017,wu2025inverse}.

The logic extends beyond low light: Gaussian noise, bicubic downsampling, and invariant blur are not universal acquisition models. Simplification is legitimate when scoped; validation should compare relevant conditional properties and use real pairs to test the simulator.

\paragraph{Benchmark mismatch in LLIE.}
LOL-v1 contains 500 captured low/normal pairs (485 training, only 15 test) \cite{wei2018retinex}; LOL-v2 adds real 689/100 and synthetic 900/100 splits \cite{yang2021lolv2}. They enable paired training, but define ``normal'' by their reference exposure and capture protocol. Limited scenes, alignment, cameras and exposure/ISO settings sample only part of illumination, dynamic range, motion, sensor, ISP and rendering diversity. LOL-v2-Synthetic further starts from an already informative normal image.

SID pairs 5,094 raw short exposures with long exposures from 424 static scenes and two cameras \cite{chen2018sid}. It captures genuine photon/read noise, but its static, sensor-specific raw-to-long-exposure task does not cover motion, processed JPEGs, other sensors, or illumination correction. SICE uses 589 multi-exposure sequences and references selected from HDR/fusion results \cite{cai2018sice}; it captures dynamic-range issues, but its target is a preferred rendering, not unique physical truth. LOL exposure translation, SID raw short-to-long imaging, and SICE tone rendering are therefore related but different conditionals.

\subsection{Low light is not one degradation variable}
The phrase ``low-light image'' conflates several causes. Global under-exposure can leave recoverable headroom; photon-limited shadows can have fundamentally poor SNR; strong illumination variation produces simultaneous deep shadow and highlight clipping; an unsuitable white balance or ISP can introduce colour error; motion can make a long-exposure reference geometrically different; and a dark artistic rendering may contain adequate sensor information but intentionally low display luminance. These conditions should not share one unqualified target.

The HDR perspective is central when bright and dark scene regions exceed the usable capture range. Enhancement is then not simply monotonic brightening. Recovering or rendering the scene requires decisions about illumination, reflectance, noise, saturation, local contrast, colour appearance, and display range. Artificially darkening a normally exposed image changes its appearance but normally preserves a transformed version of information that has already been captured. It therefore risks teaching inversion of an artificial operation rather than recovery from limited exposure and dynamic range.

\subsection{Reference ambiguity and evaluation}
For LLIE, two observers may legitimately prefer different normal renderings. Full-reference metrics penalize deviation from the chosen reference even when another output is visually or physically reasonable. Conversely, no-reference aesthetic metrics may reward brightness, contrast, or familiar colour while overlooking amplified noise or fabricated structure. Downstream recognition can establish utility for a particular machine task but not photographic fidelity.

A defensible evaluation therefore combines reference fidelity where a reference is meaningful, raw or forward consistency where acquisition is known, HDR and colour diagnostics, cross-camera transfer, human preference for rendering objectives, and task performance when the declared purpose is machine vision. Results should also be stratified by cause: uniform under-exposure, spatially varying illumination, extreme dynamic range, saturation, noise floor, motion, and processed versus raw input.

\subsection{What a Problem-First LLIE Study Would Contribute}

A problem-first LLIE study should begin by identifying which physical cause, or combination of causes, makes the observed image difficult to use. As discussed above, the two principal causes are insufficient dynamic range and low SNR. They are related but not identical. A scene containing bright and dark regions may exceed the usable dynamic range of the camera, forcing an exposure compromise between highlight saturation and unreliable shadows. A more uniformly dark scene may remain within the sensor's dynamic range but contain too few detected photons to provide reliable measurements. The first problem concerns the range of radiance that can be captured simultaneously; the second concerns the reliability of the captured signal. A study that does not distinguish these causes may improve the appearance of benchmark images without establishing which real limitation it has addressed.

The strongest contribution therefore need not be a new network backbone. It may instead be a better formulation of the LLIE problem. For example, a study could develop and validate an inverse-rendering abstraction that separates illumination, reflectance, sensor response, and display rendering sufficiently well for the intended application. Its scientific contribution would lie in showing that these variables explain important failures of existing methods and that modelling them leads to more reliable enhancement. A learned model may still be essential, but its design would follow from the proposed image-formation model rather than serve as the starting point of the research.

A second contribution could be a calibrated family of acquisition and noise models. Rather than adding arbitrary Gaussian noise to an artificially darkened normal image, such a study could model photon-dependent noise, read noise, clipping, quantization, exposure, colour response, and in-camera processing across a meaningful range of sensors. The important result would not merely be that training with more complicated degradation improves one score. It would be evidence that the model reproduces relevant properties of real low-light measurements, predicts where existing methods fail, and improves transfer to cameras or conditions not represented in the training data. A useful simulator should be judged by how faithfully it supports real-world inference, not simply by how visually convincing its synthetic images appear.

A third contribution could redefine the desired output. Many LLIE datasets treat one normally exposed photograph as the unique reference, even though enhancement may involve several different objectives. The desired result could be an estimate of scene radiance, a canonical rendering under specified illumination, an HDR representation, a visually preferred photograph, or an image optimized for a downstream vision task. These targets are not interchangeable. A representation that separates the estimated scene from its final display rendering would allow researchers to evaluate physical recovery and rendering preference separately. It would also prevent a method from being penalized simply because its valid rendering differs from one selected reference image.

Dataset design is another potentially fundamental contribution. A problem-first dataset could control variables that existing benchmarks confound, such as illumination distribution, scene dynamic range, exposure, sensor type, photon level, motion, saturation, colour processing, and target rendering. It could include separate subsets dominated by HDR limitations, low SNR, or their interaction. This would allow researchers to determine whether a method genuinely addresses a particular cause or merely performs well on the average composition of a benchmark. Carefully documenting how each input and reference was captured or constructed would make the conditional task represented by the dataset explicit.

A problem-first study may also contribute a better account of uncertainty and recoverability. Some shadow regions contain weak but recoverable measurements; others lie at or below the noise floor. Some highlights retain information in one colour channel; others are fully saturated. An enhancement method should ideally distinguish these situations. Instead of producing an equally confident output everywhere, it could estimate spatial confidence, indicate where detail is supported by the measurement, offer alternative plausible renderings, or decline to assert detail that cannot be justified. Such mechanisms are particularly important when enhanced images are used for scientific observation, medical analysis, surveillance, autonomous systems, or other applications in which plausible fabrication may be mistaken for evidence.

Evaluation should follow the claimed contribution. An inverse-rendering model should be tested for consistency with image formation and transfer across illumination conditions. A sensor/noise model should be validated against real measurements from independent devices. A proposed scene representation should be evaluated separately from the chosen display rendering. A new dataset should demonstrate which previously hidden failure modes it reveals. An uncertainty method should be assessed for calibration and its ability to identify irrecoverable regions. Conventional PSNR, SSIM, perceptual metrics, and user studies may remain useful, but no single metric can establish all these claims.

Architecture then becomes an enabling design choice. A CNN may be appropriate for efficient local processing, a transformer for modelling spatially varying illumination, a diffusion model for representing multiple plausible solutions, or a compact inverse-rendering network for interpretability and deployment efficiency. The architecture should be selected because its properties address a requirement identified by the problem analysis. Its contribution should be evaluated in terms of accuracy, robustness, computational cost, interpretability, and dependence on prior information, rather than only by whether it produces the highest score on one benchmark.

This broader allocation of scientific credit is important for the development of the field. It recognises advances in problem formulation, acquisition modelling, target definition, dataset construction, uncertainty analysis, and evaluation as substantial contributions in their own right. It also encourages researchers to investigate why low-light images are difficult rather than repeatedly optimizing models for the most convenient artificial representation of that difficulty. A problem-first LLIE study should ultimately leave the community with a clearer understanding of the problem, a more reliable way of solving or evaluating it, or new evidence about what can and cannot be recovered—not merely another entry at the top of a benchmark table.


\section{A Problem-First Research Workflow}
The framework suggests the following order:
\begin{equation}
\boxed{\mathcal H\rightarrow\mathcal I_{\rm lost/kept}\rightarrow\mathcal K
\rightarrow p(X|Y)\rightarrow\delta^*\rightarrow(F_\theta,\mathcal D).}
\label{eq:workflow}
\end{equation}
That is: (1) model acquisition; (2) identify retained and lost information; (3) state legitimate knowledge sources; (4) characterize ambiguity and uncertainty; (5) choose the estimator and loss for the intended use; and only then (6) choose architecture and training data. A physics-aware learned formulation, for example, can combine these sources:
\begin{equation}
\begin{split}
\widehat X=\arg\min_X\;&D(Y,\mathcal H(X))+\lambda_aR_a(X)\\
&+\lambda_eR_e(X)+\lambda_iR_i(X;Y).
\end{split}
\label{eq:hybrid}
\end{equation}
This is not a prescription that every method must be model-based. It is a requirement that omitted physics and imported priors be conscious, testable choices.

\subsection{Stage 1: define target, use, and acquisition}
Specify $X$, $Y$, the intended user or downstream task, and a forward-process family. Distinguish controlled variables, nuisance variables, and unknown mechanisms. Where the exact camera pipeline is unavailable, give a justified abstraction and describe what it excludes. A useful first-stage output is not necessarily one equation; it can be a causal diagram, a calibrated simulator, or a bounded family of plausible processes.

\subsection{Stage 2: audit retained and lost information}
Identify sampling limits, null spaces, clipping, quantization, noise floors, ambiguity between latent factors, and sensitivity to model error. This stage determines which desired outputs can be evidence-supported. It also reveals where collecting a better measurement may be more effective than constructing a larger estimator.

\subsection{Stage 3: inventory legitimate knowledge}
List the information sources in $\mathcal K$ and justify their relevance. External data can provide strong natural-image or semantic priors; internal recurrence can adapt to the test image; physics can constrain consistency; analytic assumptions can improve stability; and architecture can bias the solution. The inventory should also identify prohibited information, such as content-specific priors in a forensic setting or inaccessible metadata in deployment.

\subsection{Stage 4: characterize ambiguity and shift}
Describe $p(X|Y)$ conceptually even when it cannot be calculated. Ask whether it is narrow, multimodal, sensor-dependent, or preference-dependent. Define plausible shifts in $p(Y|X)$, $p(X)$, target construction, and user loss. Decide which uncertainty should be estimated and what out-of-domain behaviour is acceptable.

\subsection{Stage 5: choose the decision and evidence}
Choose the loss or utility from the application. A mean estimate, perceptual mode, posterior sample, and task-optimized rendering answer different questions. At the same time, state the evidence required for the intended claim: matched accuracy, cross-domain robustness, human preference, physical consistency, safety, latency, or calibration. Designing evaluation at this point reduces the temptation to select favourable metrics after observing results.

\subsection{Stage 6: choose data, model, and optimization}
Only after Stages 1--5 should the implementation be selected. Model capacity, pretraining, inductive bias, and compute should match the uncertainty and constraints of the problem. Training data should sample the validated forward family and relevant image distribution, while held-out tests should include purposeful shifts. This stage is where modern AI can be used most powerfully because its role is now specified.

\subsection{The workflow is iterative, not linear}
Experiments will expose deficiencies in the acquisition family or target definition. The workflow should therefore loop: failures update the problem model; new measurements reduce ambiguity; and deployment evidence expands or contracts the validated region. What it prevents is not iteration but unexamined inheritance---the assumption that an established dataset, loss, and architecture collectively define the scientific problem.

\section{From Workflow to Research Practice}
The preceding workflow is not only a conceptual ordering; it changes how the two case studies should be formulated, solved, and evaluated. The essential contrast is between starting with a model family and asking how well it performs on an established benchmark, and starting with the deployed information problem and asking which model, data, and objective it requires. The former sequence can encourage repeated substitution of the newest architecture into an inherited pipeline, followed by extensive tuning for a narrow score. The latter makes the scientific object, assumptions, and evidence explicit before optimization begins. Recent deep models contain major advances in representation and computation, but the workflow helps determine precisely what problem those advances solve and whether a reported gain survives outside the conditions under which it was selected.

\subsection{Super-Resolution Reconstructed from the Workflow}
\emph{Stage 1: model acquisition.} Rather than assume bicubic downsampling, specify a family $\mathcal H_{\rm SR}$ containing plausible optics, pixel integration, sensor sampling, aliasing, mosaicing, demosaicing, noise, compression, sharpening, and spatially varying processing. Known parameters should be calibrated; unknown parameters should be estimated or represented by a distribution. The first design question is whether the intended resolution requires additional evidence, such as subpixel-shifted burst frames, rather than a more powerful single-image prior. \emph{Stage 2: identify information loss.} Determine the passband, aliased components, null space, saturation, and noise-limited frequencies. This analysis states which detail is measurement-supported, which can become observable through complementary acquisition, and which cannot be uniquely recovered.

\emph{Stages 3--4: identify priors and ambiguity.} State whether missing detail comes from analytic regularity, external HR images, internal recurrence, semantic foundation-model knowledge, or architecture bias. Evaluate how strongly the output changes when these priors or degradation assumptions change. \emph{Stage 5: choose the decision rule.} MSE is appropriate for a conditional mean; perceptual or adversarial objectives select visually preferred solutions; diffusion posterior sampling represents multiple plausible HR images. The choice should follow the use case, not leaderboard convention. \emph{Stage 6: choose architecture and data.} Only now decide whether a CNN, transformer, unfolding network, GAN, or diffusion model is the most effective implementation, and construct training pairs from the validated acquisition family.

This order differs from much benchmark SR, where a fixed degradation and loss are inherited before architecture design. SRCNN and SwinIR demonstrate progressively more capable mappings on synthetic benchmark distributions \cite{dong2014,liang2021}; Real-ESRGAN responds with a richer high-order synthetic degradation process \cite{wang2021realesrgan}; RealSR and DRealSR replace synthetic LR with registered captures \cite{cai2019realsr,wei2020drealsr}; and diffusion restoration adds powerful priors and posterior sampling \cite{kawar2022}. By contrast, burst methods change the evidence by acquiring complementary samples \cite{wronski2019,bhat2021burst}, while null-space and unfolding methods make the information decomposition or measurement model constrain the estimator \cite{wang2023null,zhang2020usrnet}. Each changes a different workflow component. A problem-first comparison should not pool their scores as though they estimate one $p(X|Y)$: it should state whether the advance changes acquisition, the assumed forward process, the observable reconstruction, the prior-supported null space, or only the implementation, and whether the target is bicubic SR, capture-rig transfer, blind camera SR, or plausible generative upsampling.

A convincing future SR study should therefore report: matched benchmark performance; robustness across calibrated degradation families; results on independently captured real LR/HR pairs where possible; data-consistency or re-degradation error; sensitivity to blur and noise assumptions; and either uncertainty or multiple solutions in strongly ambiguous regions. Such evidence distinguishes improvement in the estimator from increased confidence in a learned prior.

\subsection{Low-Light Enhancement Reconstructed from the Workflow}
\emph{Stage 1} should begin with illumination, reflectance, scene radiance, exposure, sensor noise, clipping, quantization, colour processing, and display rendering, rather than with a gamma-darkened normal image. \emph{Stage 2} then identifies the distinct losses: photon-limited shadow information, clipped highlights, colour uncertainty, spatially nonuniform illumination, and compression or processing artifacts. This reveals why low light is simultaneously a denoising, inverse-rendering, colour, and high-dynamic-range problem.

At \emph{Stages 3--4}, the method should declare whether it assumes Retinex-type decomposition, learned normal-light statistics, semantic plausibility, repeated structure, or a calibrated camera pipeline. The inverse-rendering view in \cite{wu2025inverse} uses a simplified scene representation and canonical re-rendering, thereby changing the formulation from ``invert an artificial darkening curve'' to ``estimate and correct the causes of image formation.'' Ambiguity must still be acknowledged: saturated highlights and noise-floor shadows do not have unique inverses.

At \emph{Stage 5}, the desired output must be defined. It may be a faithful estimate of a canonical rendering, a visually pleasing photograph, an HDR representation, or an image optimized for downstream recognition. These objectives imply different losses and acceptable uses of prior-supported detail. At \emph{Stage 6}, architecture follows: a transformer may supply long-range illumination context, a diffusion model may represent multimodal uncertainty, and a compact inverse-rendering network may provide a more interpretable and data-efficient solution.

Recent LLIE research illustrates both progress and the remaining risk. Retinexformer combines an illumination model with a transformer \cite{cai2023retinexformer}; HVI improves colour and intensity representation \cite{yan2025hvi}; and challenge systems combine high-capacity models and optimize benchmark quality \cite{liu2024ntire}. Yet a model trained and tested on LOL estimates the LOL paired-exposure conditional, a SID model estimates a sensor-specific raw short-to-long exposure conditional, and SICE supplies a selected fusion rendering. High scores across rows of a benchmark table do not erase these distinctions. Under the workflow, researchers should report within-benchmark results, cross-dataset transfer (e.g., LOL-v1$\leftrightarrow$LOL-v2-Real and synthetic$\rightarrow$real), cross-sensor raw testing for SID-type methods, and evaluation on dynamic scenes and independently captured real images. Physical or decomposition consistency and downstream vision should complement PSNR, SSIM and perceptual quality.

\subsection{Guidelines for Future Research}
The two cases suggest a reusable protocol:
\begin{enumerate}
\item \textbf{Define the deployed task before the benchmark.} State the scene, sensor, processing pipeline, output representation, and intended use.
\item \textbf{Write an observation model and its uncertainty.} Distinguish known physics, estimated parameters, nuisance variables, and deliberately omitted effects.
\item \textbf{Audit information loss.} Identify null spaces, clipping, aliasing, quantization, noise floors, and other many-to-one operations.
\item \textbf{Declare every source of missing information.} Separate measurement evidence from analytic, internal, external, semantic, and architectural priors.
\item \textbf{Match the estimator to the use.} Explain whether the output is a conditional mean, mode, perceptual decision, or posterior sample, and why.
\item \textbf{Validate data and audit the benchmark.} Document reference construction and changed latent variables; compare simulated and real conditionals; reserve independent real tests and state excluded conditions.
\item \textbf{Evaluate beyond the matched benchmark.} Include cross-dataset transfer, parameter shift, sensor or pipeline shift, re-degradation consistency, prior sensitivity, uncertainty, and task-relevant outcomes.
\item \textbf{Locate novelty accurately.} Distinguish advances in problem formulation, information source, objective, representation, inference, and architecture.
\item \textbf{Make comparative claims reproducible.} Release or fully specify preprocessing, training and selection procedures, hyperparameters, compute, and evaluation code; report repeated-run variation where it can affect the ranking; and compare under matched data, supervision, parameter, and computational conditions.
\item \textbf{Treat SOTA as a scoped result, not a scientific conclusion.} State the dataset, protocol, metric, resource budget, and statistical uncertainty under which superiority holds, then separately explain what new understanding or deployment capability the result establishes.
\end{enumerate}

\subsection{Progress Under Irreducible Mismatch}
Mismatch is not an anomaly that a sufficiently large dataset will finally remove. Cameras, optics, compression pipelines, illumination, content, user preference, and deployment objectives continue to change; even a captured ``real-world'' dataset is a finite sample from one protocol. The realistic goal is therefore not distribution identity, but controlled uncertainty about the gap and demonstrable performance over a useful region of possible worlds.

First, replace a single assumed forward model by a \emph{family} $\{\mathcal H_\eta:\eta\in\Omega\}$ whose parameters describe plausible blur, noise, sampling, exposure, response, clipping, and processing. Training should cover physically meaningful combinations rather than arbitrary augmentation, and evaluation should include held-out combinations and deliberately shifted domains. Conceptually, a robust estimator minimizes risk over an uncertainty set $\mathcal Q$ around the available distributions,
\begin{equation*}
\min_F\;\sup_{q\in\mathcal Q}\;\mathbb E_q[L(F(Y),X)],
\end{equation*}
or reports both average and worst-group performance rather than only the mean on one benchmark. The purpose is not to guarantee every possible deployment, but to state the region for which evidence exists.

Second, design systems to \emph{diagnose and adapt}. At test time, degradation estimation, metadata, self-supervision, internal image statistics, or a small amount of deployment data can update the forward model or estimator. Adaptation must be constrained by measurement consistency so that a model does not merely become more confident in its prior. When the input lies outside the validated region, the system should expose low confidence, multiple plausible outputs, or a refusal to assert unsupported detail.

Third, replace the single leaderboard by a \emph{benchmark portfolio}: matched synthetic tests for controlled comparison; captured paired tests for specific devices and protocols; cross-dataset and cross-sensor transfer; unpaired in-the-wild inputs; physical consistency tests; and downstream-task or human evaluation appropriate to the application. No component is sufficient alone. A leaderboard remains useful as a controlled instrument, but the best-tuned entry is not automatically the best explanation, the most reproducible method, or the most reliable deployed system. Progress is credible when a method improves the intended target without a hidden collapse under these complementary tests, and when the improvement remains meaningful under transparent, matched experimental conditions.

Finally, treat deployment as part of the scientific loop. Record failure modes, shifts in sensors or content, and calibration error; add representative cases to evaluation without discarding earlier tests; and distinguish a general method improvement from adaptation to a newly expanded benchmark. Claims should be scoped---for example, ``robust across the tested camera/degradation family'' rather than ``solves real-world SR.'' In this way, irreducible mismatch becomes manageable: not by pretending that train and real distributions coincide, but through explicit model uncertainty, stress testing, adaptation, calibrated outputs, and cumulative evidence.

\section{Implications for Research Training and Publication}
\subsection{How students should enter the field}
A new researcher should not be required to master every classical method before using a neural network, nor should education delay engagement with current technology. The aim is integration rather than nostalgia. Researchers should learn the enduring questions alongside modern tools: image formation, sampling, convolution, noise, dynamic range, colour, ill-posedness, regularization, statistical estimation, uncertainty, decision objectives, and evaluation. Historical methods are valuable not as museum pieces but because their assumptions are visible. They demonstrate how a solution follows from a problem description and how failure follows when assumptions break.

We propose a problem-first training sequence:
\begin{enumerate}
\item \textbf{Observe the physical process.} Examine how optics, illumination, exposure, sensing, sampling, noise, and processing form the image used by the algorithm.
\item \textbf{Analyse the inference problem.} Identify retained information, lost information, confounded variables, non-identifiability, and sensitivity to model error.
\item \textbf{Reconstruct a transparent baseline.} Implement or reproduce a classical or model-based approach whose assumptions can be inspected.
\item \textbf{Interrogate the benchmark.} Reconstruct how inputs and references were created, what conditional task they define, and which real conditions they exclude.
\item \textbf{Interpret a modern model.} Trace its output to measurement evidence, physics, analytic assumptions, internal statistics, external data, and architecture bias.
\item \textbf{Create and diagnose mismatch.} Change one acquisition, sensor, content, or target factor and study why performance changes rather than merely recording that it changes.
\item \textbf{Design the solution from the diagnosis.} Select a CNN, transformer, diffusion model, hybrid method, or different measurement only after identifying the limitation that the choice is intended to overcome.
\item \textbf{Make a proportionate claim.} Match the conclusion to the validated conditions and report uncertainty, failure, and prior dependence where relevant.
\end{enumerate}

This sequence changes the meaning of technical competence. A capable researcher should be able not only to improve a model, but also to decide whether the model is solving the right problem; not only to use a dataset, but to explain the task its construction creates; and not only to report a metric, but to state what scientific conclusion that metric supports. Architecture research remains important, but becomes more original because it is directed toward a diagnosed limitation rather than a fashionable component.

\subsection{Responsibilities of supervisors and research leaders}
The educational change cannot be delegated to students alone. Supervisors and research leaders determine whether projects begin with a problem description or a model name. They can require every proposal to state the acquisition process, target, ambiguity, prior sources, deployment conditions, and falsifiable claim before approving architecture development. Group discussions can examine failure cases and mismatches with the same seriousness as average benchmark improvements.

Recruitment and mentoring can likewise assess whether candidates understand why a method works, what information it uses, and when it should fail---not only whether they can reproduce a current model. Industry and national laboratories have a particular interest in this distinction because deployment rapidly exposes assumptions hidden by academic benchmarks. Strong fundamentals therefore improve employability and technology transfer as well as scholarship.

At community level, tutorials, doctoral courses, challenges, and review criteria should connect modern models back to acquisition, inverse problems, statistical decisions, and evidence. The objective is not to create a gatekeeping canon or to privilege one generation of methods. It is to ensure that the next generation inherits the accumulated scientific understanding of image processing while gaining the full benefit of modern AI.

\subsection{Questions for authors}
Before submission, authors can ask:
\begin{enumerate}
\item Is the desired target defined independently of the dataset filename?
\item Which parts of the forward process are known, simulated, estimated, or ignored?
\item Which output details are measurement-supported and which are prior-supported?
\item Are baselines matched in data, supervision, compute, tuning, and test-time processing?
\item Is the claimed gain larger than relevant run-to-run or protocol variation?
\item What happens under an independently chosen dataset, sensor, or degradation shift?
\item Can another group reconstruct the data and evaluation pipeline from the report?
\item Does the conclusion state exactly what has improved and where it has been validated?
\end{enumerate}

\subsection{Questions for reviewers, editors, and benchmark organizers}
Reviewers should reward a well-supported problem or diagnostic contribution even when its headline metric is not the largest. They should distinguish absence of a leaderboard maximum from absence of novelty, and demand stronger evidence when a paper makes a broad real-world claim from a narrow benchmark. Conversely, a problem-first paper should not receive exemption from quantitative comparison; its evaluation should follow from its claim.

Editors, area chairs, and programme committees shape culture through reviewer guidance and selection criteria. They can state explicitly that novelty may reside in problem formulation, measurement, data, evaluation, diagnosis, or validated deployment---not only architecture. They can invite reviewers to assess whether evidence matches the claim rather than treating one benchmark maximum as a universal requirement. This does not lower standards; it makes the standards more scientifically discriminating.

Benchmark organizers can assist by documenting capture and target construction, preserving hidden tests, reporting uncertainty, separating tracks with different data and resource rules, and introducing shifted or refreshed evaluation. Leaderboards could show efficiency, calibration, worst-group performance, and cross-domain results alongside one average score. Such changes would make competition a better instrument of science.

\section{Toward a Problem-First Research Culture}
\subsection{Culture changes through everyday decisions}
Research culture is not changed by a manifesto alone. It changes through the questions supervisors ask, the projects institutions fund, the evidence reviewers request, the papers editors select, the dimensions leaderboards display, and the capabilities employers value. No single group can make the transition independently, and no group should be blamed for responding to existing incentives. The opportunity is collective: make the scientifically stronger path also a recognized and practical path to publication, recognition, and impact.

\begin{table*}[t]
\centering
\caption{Contributions to a problem-first image-processing culture.}
\label{tab:culture}
\small
\begin{tabular}{p{2.8cm}p{5.4cm}p{5.2cm}}
\toprule
Community actor & Constructive practice & Cultural signal\\
\midrule
Students and researchers & Begin with target, acquisition, information loss, assumptions, and use; treat architecture as a justified design decision & Understanding the problem is part of technical excellence\\
Supervisors and research leaders & Require a problem statement and falsifiable claim before model development; discuss failures and mismatch, not only average scores & Careful formulation and diagnosis are legitimate research outputs\\
Authors & Match claims to evidence; disclose resources and tuning; report robustness, uncertainty, and limitations & Credibility matters more than rhetorical breadth\\
Reviewers & Evaluate whether the method solves the stated problem and whether comparisons are fair; recognize non-architectural novelty & The highest familiar-benchmark score is not the only form of contribution\\
Editors and programme committees & Broaden scope and reviewer guidance; welcome problem, data, measurement, evaluation, and diagnostic papers & Publication rewards cumulative knowledge, not only leaderboard movement\\
Benchmark organizers & Document construction; refresh tests; include shifts, efficiency, uncertainty, and worst-group results & Benchmarks are instruments for inquiry rather than definitions of reality\\
Institutions and funders & Support data collection, replication, long-horizon problem study, shared infrastructure, and negative findings & Slow foundational work is a valued investment\\
Industry and laboratories & Share deployment failure modes where possible; value physical and inferential understanding in recruitment & Real-world reliability requires more than benchmark fluency\\
\bottomrule
\end{tabular}
\end{table*}

\subsection{Broadening novelty without lowering standards}
A frequent concern is that broadening the definition of contribution could make evaluation subjective or excuse weak performance. The opposite should be the aim. A problem-formulation paper must show that the existing formulation is inadequate and that the new one changes understanding or capability. A dataset paper must establish what new variation it measures and document its construction. A diagnostic paper must isolate causes and support reproducible conclusions. A robustness paper must define meaningful shifts. An architecture paper must demonstrate that its design, rather than unmatched resources, produces the claimed gain.

Thus, different contributions require different but equally rigorous evidence. The unifying standard is the alignment of problem, claim, and evidence. This is more demanding than requiring every paper to compete on the same table, because it asks reviewers to examine what has actually been learned.

\subsection{Changing the language of progress}
Language helps create expectations. Instead of saying that a method ``solves real-world super-resolution,'' authors can state the tested camera and degradation family. Instead of calling a benchmark result simply ``SOTA,'' they can specify the dataset, protocol, resources, and uncertainty under which it is superior. Instead of describing generated detail as recovered, they can distinguish measurement-supported and prior-supported content. Such qualification does not weaken a paper. It makes its contribution more durable because later work can identify precisely what has been established and what remains open.

The community can likewise celebrate multiple forms of progress: a better score under matched conditions; a method that transfers across sensors; a dataset that exposes a hidden mismatch; a physical model that explains failure; a calibrated system that knows when it is uncertain; or a negative result that prevents repeated error. A mature science needs all of these.

\subsection{A gradual and inclusive transition}
The proposed change should be evolutionary rather than adversarial. Existing benchmarks, challenges, and model families embody substantial collective effort and should be improved, not dismissed. Young researchers should not be disadvantaged for having trained within the prevailing system. Instead, courses, tutorials, review forms, benchmark tracks, and mentoring can progressively add problem-first questions and evidence. The transition succeeds if it expands researchers' capabilities while preserving the openness, speed, and empirical energy that modern AI has brought to image processing.

\section{Scope and Limitations}
This paper is a methodological framework rather than an empirical meta-analysis of all image-processing leaderboards. SR and LLIE were selected because both show how an artificial benchmark construction can be mistaken for the real physical problem. In SR, digitally downsampling a high-quality image can obscure insufficient spatial acquisition, aliasing, limited measurement quality, and ambiguity about the missing detail. In LLIE, artificially darkening a normally exposed image can obscure dynamic-range limitations, photon-limited SNR, and inverse-rendering ambiguity. Other areas---denoising, deblurring, dehazing, compression restoration, computational photography, medical imaging, remote sensing, and inverse graphics---require domain-specific acquisition and decision analysis.

The discussion of research culture is similarly a reasoned position informed by recurring professional experience, not a sociological survey of the community. Practices vary across subfields, institutions, and laboratories, and many researchers already combine strong fundamentals with modern learning. The purpose is to articulate a recognizable structural risk and propose constructive changes that can be debated, tested, and refined.

Nor do explicit physics and broad testing eliminate mismatch. Forward models remain approximations, human preference changes, and deployment can reveal unforeseen variables. Uncertainty estimates may themselves be miscalibrated under shift. The framework therefore cannot guarantee universal validity. Its aim is to make the chain of inference inspectable, the evidence proportional to the claim, and future correction cumulative.

The framework also does not prescribe one balance between fidelity and generation. A creative application may legitimately prefer a plausible, aesthetically strong image to a faithful reconstruction. A scientific application may prohibit unsupported detail. Both can be excellent image processing when the target and evidence are stated; confusion arises when success under one objective is claimed as success under the other.

\section{Conclusion}
The protocol strengthens, rather than opposes, modern AI: architecture and scale follow a defined inferential target. Its broader message is that the future of the field depends on how future researchers are educated. They should learn image processing from the problem outward---image formation, information loss, assumptions, ambiguity, and decision---rather than from the currently fashionable model inward. This requires patience: modelling acquisition, constructing meaningful data, analysing failure, and validating transfer may take longer than fine-tuning another architecture, but these are often the activities through which durable scientific progress occurs.

Across SR and LLIE, the central lesson is the same. An artificial transformation can create an image with a similar appearance without reproducing the physical process that caused the real problem. Digitally downsampling a high-quality image is not the same as failing to capture sufficient spatial detail through a real optical--sensor system. Artificially darkening a normally exposed image is not the same as recording a scene with insufficient dynamic range and low SNR. Reliable advances therefore require validated acquisition models, clearly identified priors, appropriate objectives, reproducible comparisons, cross-domain evidence, adaptation, and honest reporting of uncertainty. A benchmark gain is valuable evidence within its stated conditions. Cumulative technical advancement requires the larger step of showing that the gain reflects better acquisition, better problem formulation, justified use of information, more reliable inference, or demonstrable capability on the intended imaging problem.

Rethinking image processing is therefore a scientific, educational, and cultural project. The community must continue to develop powerful models, but it must also develop researchers able to question the tasks those models inherit, understand the physical and statistical foundations of the problem, and design evidence capable of distinguishing benchmark optimization from real advancement. Achieving this does not require rejecting current practice or assigning blame. It requires researchers, supervisors, reviewers, editors, benchmark organizers, institutions, and industry to broaden what they recognize and reward as progress.

The desired culture combines the best qualities of modern AI research---openness, rapid experimentation, shared code, empirical testing, and powerful learned models---with the enduring disciplines of image science: careful observation, physical modelling, identifiability analysis, justified assumptions, uncertainty, falsifiable claims, and patient validation. The central question should not be only ``which model wins?'' but first ``what is the problem, what information can solve it, and what evidence would show that we have genuinely advanced?''

 \section*{Acknowledgment}
 OpenAI’s ChatGPT was used during the preparation of this manuscript to assist with language revision, manuscript organisation, drafting and refining alternative formulations, and literature-search support throughout the paper. The author developed the central ideas, scientific arguments, interpretations, and conclusions; independently reviewed and revised all AI-assisted text; verified the technical content and cited sources; and takes full responsibility for the final manuscript.

\ifieee\bibliographystyle{IEEEtran}\else\bibliographystyle{unsrt}\fi
\bibliography{references}
\end{document}